\documentclass{fundam}

\usepackage{graphicx}
\usepackage{makeidx}
\usepackage{latexsym} 
\usepackage{url}

\newtheorem{Remark}{\sc Remark}[section] 
\newtheorem{Theorem}{\sc Theorem}[section]
\newtheorem{Corollary}[Theorem]{\sc Corollary}

\newtheorem{Definition}{\sc Definition}[section]

\newtheorem{Conjecture}[Theorem]{\sc Conjecture}
\newtheorem{Example}{\sc Example}[section]

\begin{document}

\setcounter{page}{1}


\title{Evolutionary Automata and Deep Evolutionary Computation}

\author{Eugene Eberbach\thanks{Retired Professor of Practice (Rensselaer Polytechnic Institute); Corresponding author: eeberbach@gmail.com (Eugene Eberbach)  \newline\newline
                    \vspace*{-6mm}{\scriptsize{Received November 2024; modified August 2025 
}}}
\\
Dept. of Eng. and Science,
Rensselaer Polytechnic Institute\\ 275 Windsor Street, Hartford, CT 06120, USA
}
\maketitle

\runninghead{E. Eberbach}{Evolutionary Automata}

\vspace*{-6mm}
   


\begin{abstract}
Evolution by natural selection, which is one of the most compelling themes of modern science, brought forth evolutionary algorithms and evolutionary computation, applying mechanisms of evolution in nature to various problems solved by computers. In this paper we concentrate on evolutionary automata that constitute an analogous model of evolutionary computation compared to well-known evolutionary algorithms. Evolutionary automata provide a more complete dual model of evolutionary computation, similarly like abstract automata (e.g., Turing machines) form a more formal and precise model compared to recursive algorithms and their subset - evolutionary algorithms. An evolutionary automaton is an automaton that evolves performing evolutionary computation perhaps using an infinite number of generations. This model allows for a direct modeling evolution of evolution, and leads to tremendous expressiveness of   evolutionary automata and evolutionary computation. This also gives the hint to the power of natural evolution that is self-evolving by interactive feedback with the environment.

\end{abstract}


\section{Introduction}

The research motivation of this paper is to make evolutionary computation more attractive for a general reader, similar like deep neural networks revived neural networks research and applications in 2006. Additionally, we want to strengthen the very incomplete foundations of evolutionary computation, concentrating mostly on investigation of its expressiveness. Another motivation is that similarly like evolutionary computation has been inspired by natural evolution, perhaps, vice versa, natural evolution may gain something from that research too, i.e., to stir fruitful discussion, whether natural evolution is finite, self-evolving, and goal-directed. 

The main contribution of this paper is showing that evolutionary automata cover all existing and not yet invented types of evolutionary algorithms, and proving that they could solve not only intractable hard optimization problems, but Turing machine\footnote{for the definition of abstract automata (and Turing machines, in particular) the reader is advised to look at any classical textbook on foundations of computer science, e.g., \cite{hopcroft01}}  unsolvable/undecidable problems as well. For the first time, the power of evolutionary automata with other super-Turing models of computation is compared. Assuming that features of evolutionary automata could be extrapolated back to natural evolution as well, the hint why natural evolution is so powerful is presented. Finally, due to introduced new concepts, the strikingly simple proof of the convergence in the limit of evolutionary algorithms is given.

{\em As it is well known, in evolutionary computation (EC), selection operates on population of individuals that are evolved using mutation and crossover in much the same way as natural evolution does. EC can be understood as a probabilistic beam search based on principles of natural evolution and directed by fitness performance measure to solve hard optimization problems.}

Evolutionary computation consists of four main areas: {\em Genetic Algorithms} (GA) \cite{holland75}, {\em Genetic Programming} (GP) \cite{koza92}, {\em Evolution Strategies} (ES) \cite{rech73} and {\em Evolutionary Programming} (EP) \cite{fogel66}. Sometimes some authors question whether GP deserves to be included in the four main categories. We are convinced that it does.

Additional areas include: Ant Colony Optimization (ACO), Particle Swarm Optimization (PSO), co-evolution, Artificial Immune Systems (AIS), evolutionary robotics, evolvable hardware, Evolutionary Artificial Neural Networks (EANN), evolutionary multi-objective optimization, Artificial Life (A-Life), Classifier Systems, DNA-Based Computing, Evolutionary Bioinformatics, Differential Evolution, Bacterial Foraging, Artificial Bees, Fireflies Algorithm, Harmony Search, Digital Organisms, Stochastic Diffusion Search, Nano Computing, Membrane Computing, Human-centric Computing, Memetic Computing, Autonomic Computing and  Self-organizing Systems (see, e.g. \cite{back97,fogel95,michalewicz96,michfog04}).

Evolutionary automata, compared to evolutionary algorithms, provide a more complete, dual and more general model of evolutionary computing, similarly like abstract automata (e.g., Turing machines) form a more formal and precise model compared to recursive algorithms and their subset - evolutionary algorithms. 

This research, although based on previous publications of the author and many other scientists, is completely new and original. The author is not aware about a similar approach to evolutionary computing. Probably, the closest in the spirit to our approach is the book by K. A.  De Jong, Evolutionary Algorithms: A Unified Approach, The MIT Press, 2006. However, this unification from the above book is about classical finite evolutionary algorithms in the style of our section 2 on Evolutionary Algorithms. It does not cover the way to solve global optima in infinity, nor TM undecidable problems.  In our approach, due to introduced new concepts, many difficult topics (e.g., undecidability, infinite computations, convergence in the limit, self-evolution) become much simpler and almost obvious. Note that evolutionary automata consist of abstract automata of the type of finite automata, pushdown automata, linearly bounded automata, Turing machines, and so on, and they may evolve in successive generations. For example, finite automata are already so general, that they include practically all known types of evolutionary algorithms as well as swarm intelligence algorithms, because both subareas consist typically of simple iterative algorithms of the power of regular languages, and these are exactly evolutionary finite automata - the simplest kind of evolutionary automata from this paper.

Note that challenges of this research are very big. Firstly, it is very difficult to convince scientific community to a new approach. Secondly, we open truly the Pandora's box, whether everything in the ``real'' world is finite or not. Thirdly, whether to solve Turing machine unsolvable problems is possible, and if so, in which sense? Many researchers consider still such research with skepticism similar like was with the AI research before 1980s and launching the {\bf Fifth Generation Computer Project} in Japan, Europe, and USA \cite{refenes89,tohru85}. We are strongly convinced that such research is badly needed, because computer science will have to tackle and solve such problems in forthcoming near future.

Problems can be either solvable or unsolvable (called also undecidable) using a specific model/theory. In computer science, Turing machines form such dominating and most popular model for problem solving. Some problems are Turing machine solvable and some not. Our paper intends to show that evolutionary computation, besides decidable hard computational problems, may deal with Turing machine undecidable problems too. The typical belief is that proving that a specific problem is TM-undecidable stops any attempt to solve that problem and that is the end of the story. On the other hand, we are convinced that this is only the beginning. Firstly, we may decide special instances (or perhaps even almost all instances) of the undecidable problem. For example, if we have the probability distribution of input instances, perhaps randomized techniques may help to estimate for which inputs problems are decidable. Secondly, we can approximate the solutions and we may decide the specific instances either in a finite number of steps or asymptotically in the infinity. There are other approaches possible to deal with undecidability too (see, e.g., the infinity, evolution, and interaction principles from \cite{ebeweg03,ebegolweg04}), and all above looks like an excellent and exiting new venue for many years of fruitful research to come.

The justification why we need new generation computing and new models of computation are the following. In \cite{denning11}, Peter Denning expressed concerns that developments in non-terminating computation, analog computation, continuous computation, and natural computation may require rethinking the basic definitions of computation. He further stated that computation is the process that the machine or algorithm generates. In analogies: the machine is a car, the desired outcome is the driver's destination, and the computation is the journey taken by the car and driver to the destination. 

On the other hand, Alfred Aho, wrote in \cite{aho11} ``as the computer systems we wish to build become more complex and as we apply computer science abstractions to new domains, we discover that we do not always have the appropriate models to devise solutions. In these cases, computational thinking becomes a research activity that includes inventing appropriate new models of computation''. It looks that the creation of new models of computation will be a never-ending story. Unfortunately, the elegant Turing machines do not constitute the final and complete model of computation (despite that many computer scientists and the author of this paper would like to be so).

As was written in \cite{wegebebur12}: 

\begin{quote}
{\em Turing Machines can be considered as an attempt to create the theory of everything for computer science, whereas similar attempts of complete theories for physics (by Newton, Laplace, Einstein, Hawking), mathematics (by Hilbert, G\"{o}del, Church, Turing) or philosophy (by Aristotle, Plato, Hegel) have failed. If Turing machines were truly complete, computer science with its Turing machine model would be an exception from other sciences, and computer science together with its Turing machine model would be complete. If so, by reduction techniques, we could prove also completeness of mathematics (decision problem in mathematics \cite{whitehead12} - disproved by G\"{o}del, Church, and Turing \cite{godel31,church36,turing36}), and completeness of physics, philosophy, medicine, economy and so on.}
\end{quote}

This paper can be considered in some sense as the follow up of \cite{ebe21}, with the restriction that we concentrate on one specific super-Turing model of computations only, called evolutionary automata that is applied to investigate the expressiveness and other properties of evolutionary computing.

Natural evolution does not use an explicit fitness function. However, it indirectly, by interaction with the environment, promotes fitter individuals leading to change of crossover and mutation rates.

Additionally, it looks that natural evolution does not have to have a limited time span; even if we strongly believe in the scientific unproven hypothesis that the Sun will no longer support life on the Earth, because it will explode after 5 billion of years. However, if this is so, we do not know whether this will apply to the whole Universe as well, i.e., whether life and natural evolution is restricted to the Earth only. Scientists claim that it is very unlikely that the biological life, as we know it, is restricted to our planet only, and, perhaps, human race will be able to escape somehow to other inhabitable parts of the Universe before hypothetical  Sun's explosion.  

Assuming further that it might be possible to assume that nothing in nature is infinite (but this is only a reasonable scientific hypothesis that may be true or false), does this apply also to models of computation like Turing Machines TM using an infinite tape, pushdown automata with an infinite stack, convergent or not convergent infinite arithmetic or geometric series, infinite summation in definite integrals, infinity in probability distribution, infinity of natural or real numbers? In other words, will mathematics cease to exist after Sun explosion?  We do not know the definite answers to those crucial philosophical questions, and we can safely assume that mathematical models of computation, and, artificial evolution, in particular, can and will use safely infinity as they did so far, because that abstraction has been proven to be beneficial for humanity, and is orthogonal to the biological life's length.

Currently, artificial evolutionary computation uses mostly (with some exceptions) fixed evolutionary algorithms, an explicit fitness function, and is restricted to a finite number of generations only. Because artificial evolution is a mathematical abstraction, thus it does not need to be restricted to finiteness only. Of course, like in approximation algorithms, an infinity has to be and is approximated in real applications.

In this paper, we argue that the above restrictions lead, by necessity, to reduced expressiveness of EC. To investigate that more precisely, we use the evolutionary automata model. In that model, evolutionary algorithms may evolve too, and the number of generations can be extended to infinity. By analogy to deep neural networks, we call this type of EC - a {\bf deep evolutionary computation}.  This is in  analogy to deep neural networks with the increased number of layers.  A simple extension of the number of generations in EC, corresponding to NN layers, to infinity, leads to tremendous expressiveness of deep EC that is able to tackle Turing Machine (TM) undecidable problems, and this is dissimilar to conventional evolutionary algorithms that are restricted to polynomial and intractable problems only. Note that if to allow in deep neural networks to increase the number of layers to infinity, such deep neural networks could solve TM undecidable problems too. 

{\em The goal of this paper is to show that there are more possibilities for expanding what can be done with evolutionary algorithms, rather than a step toward a practical guide as to how to accomplish that. That, hopefully, will happen in the future.}

This paper is organized as follows. In section 2, we describe evolutionary algorithms. Section 3 outlines selected topics from the theory of evolutionary computation. In section 4, we introduce the evolutionary automata model, and section 5 deals with its expressiveness.  In section 6 we compare expressiveness of evolutionary automata versus other models of computation. Section 7 deals with computational complexity of evolutionary automata. Finally, section 8 contains the conclusions and final comments.

\section{Evolutionary Algorithms}

Evolutionary computations are directed by evolutionary algorithms. In technical terms, an evolutionary algorithm is a probabilistic beam hill climbing search algorithm directed by the chosen fitness function. It means that the beam (population size) maintains multiple search points, hill climbing implies that only a current search point from the search tree is remembered and used for optimization (going to the top of the hill), and the termination condition very often is set to the optimum of the fitness function.

\begin{Remark}  From the above, it is obvious that evolutionary algorithms, by performing hill-climbing, can be classified as {\em local search algorithms}. However, they can be classified also as {\em approximation algorithms}  (they typically approximate only global optima for intractable problems), or {\em randomized algorithms} (they typically use probabilistic variation and selection operators). They can be classified also as {\em parallel algorithms} (they run simultaneously a set of solutions). On top of that, if we allow evolutionary algorithms to use an infinite number of generations (like in natural evolution or evolutionary automata), we get algorithms that {\em run forever}. As we see the evolutionary algorithms form a strange hybrid class having features of local search, approximation, randomized, parallel and running forever algorithms. For more details, the reader is advised to look, for instance, to \cite{kleinberg06}.
\end{Remark}

Let $X$  be the {\em representation space}, also called the {\em optimization space}, for species (systems) used in the process of optimization and a {\em  fitness function}  $f: X \rightarrow R$  is chosen \cite{fogel95,fogel01,michfog04}.

\begin{Definition}[On evolutionary algorithms]
A {\em generic evolutionary algorithm} (EA) E can be represented as the collection

\begin{center}
$E = (X, X[0], F, f, s, \nu, C)$
\end{center}
and described in the form of the functional equation (recurrence relation) $r$ working in a simple iterative loop in discrete time $t$, defining generations $X[t]$, $t = 0, 1, 2, 3,...$:

\begin{center}
$X[t+1] = s( \nu (X[t]))$, where
\end{center}

\begin{itemize}
\item 
$X$ is a {\em representation space} in the form of the finite set called {\em population}; (e.g., $X$ consists of the set of fixed binary strings for genetic algorithms (GAs), Finite State Automata for evolutionary programming (EP), parse trees for genetic programming (GP), or vectors of real numbers for evolution strategies (ES));
\item 
{\em selection operators} $s_i$ (e.g., {\em truncation, proportional selection} or {\em tournament}, $i = 1, 2, 3,...;$
\item
{\em variation operators} $\nu_i$ (e.g., {\em mutation, crossover/recombination, majority logic} or some combination of mutation and crossover), $i = 1, 2, 3,...;$
\item
a {\em fitness function} $ f: X \rightarrow  R$, which typically takes values in the domain of nonnegative real numbers and is extended to the subsets of the set $X$  by the following rule                                              
      if $ Y \subseteq X$, then $ f(Y) = max \{f(x), x \in Y \}$
\item
a {\em termination or search condition} (goal of evolution) $C$;
\item
$X[0]$ is an {\em initial population};
\item
$X[t] \subseteq X$ is the population produced on the $(t-1)$-th stage of the evolutionary algorithm (EA) $A$;
\item
$F \subseteq X$ is the set of {\em final populations} satisfying the {\em termination condition}(goal of evolution). 
\end{itemize}
\end{Definition}

Often the termination condition of an evolutionary algorithm is given as a subset $F$ of the representation space $X$.  Computation halts when an element from $F$ is obtained. Another form of a termination condition is optimum (maximum or minimum) of the fitness function $f(x)$, which is extended to the fitness function $ f(X[t])$ of the best individual in the population $ X[t] \in F$, where $f(x)$ typically takes values in the domain of nonnegative real numbers. Computation, for example, halts when a maximum of the fitness function $f(x)$ is obtained. In many cases, it is impossible to achieve or verify this optimum. Thus, another termination condition is used (e.g., the maximum number of generations or the lack of progress through several generations). 

Dynamics of the evolutionary algorithm $A$ is described in the form of the functional equation (recurrence relation) working in a simple iterative loop with parts of the space $X$ called generations in discrete time $t = 0, 1, 2, 3,...$  \cite{fogel95,fogel01,michfog04}: $X[t+1] = s( \nu (X[t]))$.

This functional equation describes how the evolutionary algorithm $A$ taking the generation $X[t] \subseteq X$ produces the generation $X[t + 1] \subseteq X$. An initial population $X [0] \subseteq X$ is given as the input of the evolutionary algorithm. Selection is based on the fitness function $f(x)$, which is often extended from elements of $X$ to subsets of $X$, giving the best value on the elements in this subset as its value for this subset.

The above definition is applicable to all typical evolutionary algorithms, including GA, EP, ES, GP. It is possible to use it to describe emerging more complex variants of evolutionary algorithms.

Evolutionary algorithms evolve population of solutions $X$, but they may be the subject of self-adaptation (like in ES) as well. For sure, evolution in nature is not static, the rate of evolution fluctuates, their equivalents of ``variation operators'' are subject to slow or fast changes, and its goal (if it exists at all) can be a subject of modifications as well.

\begin{Remark} As we wrote before, artificial evolutionary computation uses mostly (with some exceptions as we mentioned above) fixed static evolutionary algorithms with an explicit fitness function, and is restricted to a finite number of generations only. These restrictions lead to situation that artificial evolution compared to natural one becomes unnecessarily less expressive (see following sections) compared to its potentials. Thus, we advocate in this paper, that evolutionary algorithms should be allowed to evolve and to use an infinite number of generations, if desirable. But the above requires the extension of classical theory of evolutionary computation as we know so far.
\end{Remark}

\section{Evolutionary Computation Theory}

Evolutionary computation theory is still very young and incomplete \cite{back97,fogel95,fogel01,michfog04}. Of course, some limited progress has been made since early 2000s, but it is not a striking progress compared to e.g., deep neural networks that revived struggling neural networks theory and practice in 2000s. Exactly, in this context evolutionary automata are our proposal to make a more substantial progress in EC as well.

For evolutionary computation, the properties studied include among others:

\begin{itemize}
\item
Convergence in the limit 
\item
Convergence rates 
\item
Building Block Analysis 
\item
Best variation operators 
\item
Scalability 
\item
Expressiveness
\end{itemize}

\subsection{Convergence in the limit}
Canonical Genetic Algorithms (EC) do not converge to globally optimal solutions in the limit \cite{rudolph94}. Without elitism (e.g., using proportional selection) EC algorithms are divergent.

Several authors proposed necessary and sufficient conditions for convergence of evolutionary algorithms in the limit. For example, Michalewicz's contractive mapping GAs \cite{michalewicz96}, and (1+1)-ES \cite{rech73}, EP with truncation \cite{fogel01} use the elitist selection, and they converge to global optimum in infinity. David Fogel's proof of convergence of EC with elitism + ergodic variation is based on Markov chain analysis \cite{fogel95,fogel01}.
Summarizing these efforts, we can write briefly the below theorem, where our contribution provides only a new very simple proof of convergence in the limit, which is possible due to new introduced concepts. Previous proofs were quite complicated and obvolute. Note, the theorem is not new, however strikingly simple proof is due to our new theoretical concepts.

\begin{Theorem}[On global convergence of EC in the limit]
Evolutionary Algorithms (i.e., GA, GP, ES, EP) with ``elitist'' selection (preserving a current best individual from generation to generation) and assuming that variation operators are complete, i.e., allow to explore (reach) any search point in the search space, will converge asymptotically to globally optimal solutions in the limit (as number of generations goes to infinity).

{\em Proof:} If search is complete, then any search point (including global optimum) is guaranteed to be reached in infinity. If variation operators allow to reach any solution, i.e., probability to reach them is nonzero, thus in infinity evolutionary algorithms will explore all solutions by an exhaustive search; elitism combined with exhaustive search (exhaustive - because the probability to reach any solution is nonzero, thus in infinity every solution will be explored) will monotonically improve solutions, and finally the best will be reached (note, for the finite number of generations, exhaustive search is not guaranteed, thus a global optimum may not be reached). Elitism will guarantee that global optimum will be maintained and not lost. $\Box$
\end{Theorem}

Note that the requirement for search completeness is very strong (see, e.g., \cite{russell95}). For example, A*, IDA*, Minimax, alpha-beta pruning are complete, but hill climbing, evolutionary algorithms, gradient descent techniques are not complete in a general case. However, random-restart hill climbing, simulated annealing and Boltzmann machine with temperature parameter ``decreased slowly'' are complete - they simulate very well exhaustive search. Note a subtle difference: EC with elitism + appropriate (this means complete, i.e., allowing to reach any solution) variation operator guarantees to find a global optimum in infinity; without elitism - a global optimum can be ``by chance'' reached, however it is not guaranteed to be preserved (because we may not know that global optimum has been reached)! Both completeness and elitism are required to reach global optimum (completeness) and to maintain it (elitism).  
                    
\subsection{Convergence rates}

Rates of convergence have been proven for strongly convex functions (due to work for ES \cite{rech73})
\begin{itemize}
\item
$(1+1)$-ES strategy gives geometric rate of convergence
\item
$(1, \lambda)$-ES accelerates by $O(log \lambda)$
\end{itemize}

How does this compare with the rest of EC?
No rates of convergence proven so far for general cases.

\subsection{Building Block Analysis}

{\em Schema theorems} \cite{holland75,poli01} describe the effect of evolution (combined crossover, mutation, and selection) on populations.
Currently, schema theorems exist for GA and GP (but not for ES or EP - schema theorems can and have to be developed separately for each specific representation, variation and selection operators) Instead of performing analysis of all population members and treat them as a dynamical system with Markov chain analysis, the search space (population) is divided into schemata (schemas), which makes the whole analysis simpler in theory.
Schema theorems describe how and why the individuals in the population move on average from one schema (subspace) to another in one generation using Markov chain analysis.

\begin{Remark}
Schema theorems do not say anything about the convergence (limit or rate) - this would require multiple/infinite generation steady-state analysis - something in the style of M/M/1 queue (nobody did it so far - enough problems with one generation ahead).
\end{Remark}

For GA - schemas are represented by strings of $0, 1$, and $\#$ (don't care), for GP - schemas are represented by trees of the same size and shape over alphabet of functions $F$ and terminals $T$, and $=$ (don't care - to replace a single terminal or function) or $\#$ (don't care - to replace any valid subtree)
Schema theorems are provided in two variants \cite{poli01}:

\begin{itemize}
\item
estimate - the ``worst-case'' pessimistic schema theorems (with lower bound limit inequality ``$\geq$'')
\item
exact schema theorems (using equal sign ``$=$'')
\end{itemize}

Note that incorrectly, schema theorems were initially associated with the whole theory of evolutionary computation for many years (see, e.g. \cite{back97,holland75,michfog04}, whereas they are rather useless for that (they do not scale well to new subareas of EC besides GA and GP, and, additionally, say almost nothing about the convergence in the limit, convergence rate, best variation operators, scalability or expressiveness of evolutionary computation).

\subsection{Best variation operators}

David Wolpert  \cite{wolpert97} proved {\em No Free Lunch (NFL) Theorem} assuming uniform problem distribution
for learning in 1992 and 1994, and
for optimization in 1995 (with Macready)

A number of ``no free lunch'' (NFL) theorems demonstrate that for any algorithm, any elevated performance over one class of problems is offset by performance over another class. The NFL Theorem in plain words means: For all algorithms that do not resample points, their measured performance will be identical when applied across all possible cost functions on average (and at each iteration)
Consequence: Hill-climbing without resampling is as effective at maximizing as blind random search without resampling on average. In other words: NFL theorem says that all approaches are the same if applied across all problems.

\begin{Remark} Note that the NFL Theorem does not state (like it is incorrectly interpreted sometimes) that the best algorithm for everything does not exist (this would be re-stating of the Halting Problem for the Universal Turing Machine).
\end{Remark}

\begin{Remark} Note, we never want to do everything at a given moment of time, i.e., we either go to the movie or do not go to the movie, thus we do not design an algorithm ``to go to the movie'' and ``to not go to the movie'' at the same time, but one version or another. And then recipe ``to go to the movie'' will be quite different from ``not to go to the movie''. Thus, for specific (not all) problems there are {\em better} and {\em worse} solutions, and {\em not all of them are equivalent}.
\end{Remark}

\subsection{Scalability}

Scalability is associated, for example, with on-line optimization and \$-calculus \cite{ebe07,ebe22} total optimization.
Typical evolutionary algorithms are off-line, i.e., first a complete evolution is performed and as the result, the complete solution is obtained. To achieve scalability, it would be desirable to interleave process of evolution with partially evolved solutions.
\$-calculus \cite{ebe07,ebe22} is the theory for automatic problem solving and automatic programming targeting intractable and undecidable problems. It allows to express arbitrary evolutionary algorithms that are not static, can evolve, leading to evolution of evolution. In particular, \$-calculus in the style of anytime algorithms allows to evolve simultaneously the quality of solutions and costs of evolutionary algorithms, thus it produces an approximate total solution with minimal resources used (e.g., time or memory requirements).  

\subsection{Expressiveness} 

We know now much more about EC expressiveness. The Evolutionary Turing Machine model has been introduced in \cite{ebe02,ebe05}, where it has been indicated that EC might be more expressive than Turing Machines, i.e., EC can be non-algorithmic, can evolve non-recursive functions, and, in particular, can solve the halting problem of the Universal Turing Machine \cite{hopcroft01,turing36}.
Evolutionary automata, a generalization of Evolutionary Turing machines, have been introduced in order to investigate more precisely properties of EC \cite{burebe10,burebe12,burebe13,ebe02}.  In particular, they allowed to obtain new results on expressiveness of EC \cite{burebe12}. It confirms the initial result: Evolutionary finite automata, the simplest subclass of Evolutionary automata working in terminal mode can accept arbitrary languages over a given alphabet, including non-RE (e.g., diagonalization language) and recursively enumerable but not recursive languages (e.g., language of the universal TM) \cite{hopcroft01}.

\section{Evolutionary Automata}

Most of the approaches to foundations of EC do not introduce automata models - rather they apply the high-quality mathematical apparatus to existing process models, such as Markov chains, etc. They also cover only some aspects of evolutionary computation like convergence or convergence rate. 
At the same time, very little has been known about expressiveness or computational power of evolutionary computation and its scalability. In other words, evolutionary computation is not treated as a distinct and complete area with its own distinct model situated in the context of general computational models. 

This means that in spite of intensive usage of mathematical techniques, evolutionary computation has incomplete theoretical foundations. 
As a result, many properties of evolutionary processes could not be precisely studied or even found by researchers. Conventional computation has many models. One of the most popular is Turing Machine. In contrast to this, until recently evolutionary computation did not have a theoretical model able to represent practice in this domain. Only recently a rigorous mathematical foundation of evolutionary computation has been created \cite{burebe09,burebe10,burebe12,burebe13,ebe02,ebe05} although it provides only the beginning of a rigorous mathematical theory of evolutionary computation. In this theory, evolutionary automata play the role similar to the role of Turing machines, finite automata and other mathematical models in the general theory of computation. 

The approach is aimed at providing more rigorous foundations for evolutionary computation. It is based on Evolutionary Turing machine (ETM) model \cite{ebe02,ebe05}, grid automata \cite{burgin03} and super-recursive algorithms \cite{burgin99,burgin05}. This approach provides flexible tools for estimating convergence and expressiveness of evolutionary processes and algorithms, as well as for developing efficient evolutionary algorithm architectures for solving problems in science and technology. Using these tools, we were able to prove that to reach an optimum in a general case, algorithmic evolutionary processes require, in general, an infinite number of steps. This goes beyond classical recursive algorithms and Turing machines. The first constructed versions of our model, sequential evolutionary Turing machine \cite{ebe02,ebe05} and weighted evolutionary Turing machine, provide a generic theoretical model for evolutionary computation in the case of mono-evolution when a single agent performs evolution of generations and a single solution for one individual is designated to represent the whole population. An evolutionary Turing machine is an extension of the conventional Turing machine, which goes beyond the Turing machine and belongs to the class of super-recursive algorithms \cite{burgin99,burgin05}. 

However, Evolutionary Turing machines form only one class in a big diversity of evolutionary automata introduced and studied in \cite{burebe09,burebe10,burebe13}. 
This, a more general model of evolutionary computation, was used to explore universality of basic evolutionary finite automata \cite{ebebur09} and expressiveness of evolutionary finite automata \cite{burebe12}. 

Let's define a formal algorithmic model of evolutionary computation - an {\em  evolutionary automaton }also called an {\em evolutionary machine} . Let { \bf K} be a class of all abstract automata. Obviously , evolutionary automata are the subclass of { \bf K} .

\begin{Definition} [On evolutionary automata]   
A {\em basic evolutionary } {\bf K}-{\em automaton} (EA), also called {\em basic  basic evolutionary}  {\bf K}-{\em machine} is a (possibly infinite) sequence $E = \{E[t]; t = 0, 1, 2, 3, ...\}$ of automata $E[t] $ from {\bf K} each working on the population $X[t] , t = 0, 1, 2, 3, ...$ , where: 
\begin{itemize}
\item 
the automaton $E[t]$ called a {\em component}, or more exactly, a {\em level automaton}, of $E$ represents (encodes) a one-level evolutionary algorithm that works with the generation $X[t]$ of the population by applying the variation operators $\nu(t) $  and selection operator $s(t)$; 
\item
the initial $0-th$ generation  $X [0]$ is given as input to $E $ and is processed by the automaton $E[0]$, which generates/produces the first generation $ X[1] $ as its output, which goes to the automaton $E[1]$; 
\item 
for all $t = 1, 2, 3, ...$, the generation $X [t + 1]$ is obtained by applying the variation operator $\nu(t)$ and selection operator $s(t)$  to the generation $X[t] $ and these operations are performed by the automaton $E[t]$, which receives $X[t]$ as its input; the goal of the EA $E$ is to build a population $Z$ satisfying the search condition.
\end{itemize}
\end{Definition} 

The desirable search condition is the optimum of the fitness performance measure $f(x[t])$ of the best individual from the population $X[t]$. There are different modes of the EA functioning and different termination strategies. When the search condition is satisfied, then working in the recursive mode, the EA $E$  halts ($t$ stops to be incremented), otherwise a new input population $X[t + 1]$ is generated by $E[t]$. In the inductive mode, it is not necessary to halt to give the result [6]. When the search condition is satisfied and $E$ is working in the inductive mode, the EA $E$ stabilizes (the population $X[t]$ stops changing), otherwise a new input population $X[t + 1]$ is generated.

To illustrate work of evolutionary automata, a general flowchart is presented.  

\begin{figure}[!h]
\vspace*{-2cm} 
\begin{center}
\includegraphics[width=15cm]{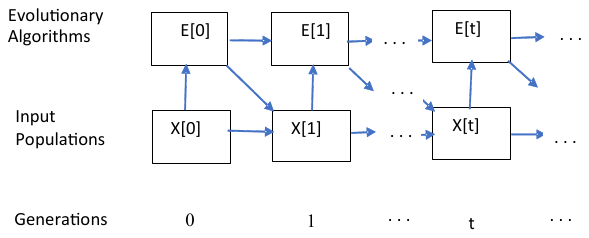}
\end{center}
\vspace*{-11cm}
\caption{Evolutionary Automaton}
\end{figure}

\vspace{1cm}
\begin{Remark}
 Note that our approach is inspired, but not equivalent to Lary Fogel's EP \cite{fogel66}. EP uses iterative evolutionary algorithms of the class of finite automata that operate on the set of finite automata $X$.  On the other hand, evolutionary automata can use evolutionary algorithms from the class of arbitrary abstract automata, e.g., finite automata, pushdown automata, linearly bounded automata, Turing machines, and so on. Also, the population space $X $ can be arbitrary, e.g., fixed binary strings, vectors of real numbers, finite automata, parse trees, etc.  However, surprisingly, despite of those restrictions Lary Fogel's Evolutionary Programming EP, if to allow an infinite number of generations and self-evolution of evolutionary algorithms, is equally expressive as other classes of evolutionary automata, what we stress in this paper and the first time we suggested in \cite{ebe05}.
\end{Remark}

Thus, Evolutionary finite automata are an important and simplest class of evolutionary machines \cite{burebe09}.

\begin{Definition} [On Evolutionary finite automata]
An {\em Evolutionary finite automaton (EFA)} is an evolutionary machine $E$ in which all automata $E[t]$ are finite automata $G[t]$ each working on the population $X[t]$ in generations $t = 0, 1, 2, 3, ... $.
\end{Definition} 

We denote the class of all evolutionary finite automata by {\bf EFA}.

It is possible to consider deterministic finite automata, which form the class {\bf DFA}, and nondeterministic finite automata, which form the class {\bf NFA}. This gives us two classes of evolutionary finite automata: {\bf EDFA} of all deterministic evolutionary finite automata and {\bf ENFA} of all nondeterministic evolutionary finite automata. Note that it is also possible to consider reconfigurable evolutionary finite automata.

\begin{Remark}
Note that all classical evolutionary algorithms GA, GP, ES, EP are fixed finite automata $E[t]$, i.e., they belong to  {\bf EFA} class -  represented by simple iterative algorithms restricted to a finite number of generations only. Note, it does not matter how obvolute and complex are user defined fitness function, or new variation or selection operators - the algorithms are computed in a simple iterative loop (feature of regular languages or finite state automata). We can describe the behavior of evolutionary algorithms as the following regular expression: $initialization \;\;(variation\; fitness-eval\; selection\; goal-not-met)^*\;goal-met$.  Evolutionary algorithms differ mostly in representation of evolved input populations  $X(i)$ - as fixed binary strings for GA, parsing trees for GP, vectors of reals for ES, and finite automata for EP. There are small and not substantial differences in variation and selection operators.  Note that only for ES  $E[t]$ evolve, i.e., the standard deviation in mutation is modified. Similarly, the differences, for example, in PSO or ASO from GA, GP, ES, EP are superficial. Of course, swarm algorithms represent different evolutionary algorithms, which for some niche applications perform better or worse than four basic EC classes, but in essence, they are finite automata meta-algorithms from the  {\b EFA} class too. 
\end{Remark}

\begin{Remark}
Note that DFA components either stop if input - set of strings are accepted or transfer their non-accepted strings to the next generation component. Technically, we can call such DFAs as Meale/Moore finite automata with outputs, or transducers.
\end{Remark}

\begin{Remark} Note that we do not specify in the definition of evolutionary automata how to evolve evolutionary machines $E[t], t=0, 1, 2, 3,... $ from generation to generation. We do this INTENTIONALLY to keep that flexible and open, and allow users to experiment with various approaches. This can be done for example using traditional mutation and crossover operators (like in evolution strategies and that can take many generations by trials and errors), or to speed up the whole process and to use the background knowledge/reverse engineering if we know which string should be accepted in the next generation. For example, we can mutate like in EP a finite state machine to obtain FSM accepting next needed word like in proofs of Theorems 5.1 and 5.2 from the next section. Intentionally, we left that open for practitioners to allow to experiment and not restrict them to one approach only (similarly like evolutionary algorithms do not enforce one type of crossover, mutation or selection operators. Cellular automata and Turing machine do not enforce one type of transition relation either). Additionally, that is not needed for the proof of expressiveness of evolutionary automata.
\end{Remark}

Evolutionary Turing machines (ETM) \cite{ebe02,burebe13} are another important class of evolutionary machines.

\begin{Definition}[On Evolutionary Turing machines] 
An {\em Evolutionary Turing machine (ETM)}  $E = \{TM[t]; t = 0, 1, 2, 3,...\}$  is an evolutionary machine $E$ in which all automata $E[t]$ are Turing machines $TM[t] $ each working on populations $X[t]$ in generations $t = 0, 1, 2, 3,...$.
\end{Definition}  

Turing machines $TM[t]$ as components of the ETM $E $ perform multiple computations in the sense of \cite{burgin05}.

\begin{Definition} [On Evolutionary inductive Turing machines] 
A {\em basic Evolutionary inductive Turing machine (EITM)} 
$EI = \{ITM[t]; t = 0, 1, 2,3 ...\}$  is an evolutionary machine $E$ in which all level automata $E[t]$ are inductive Turing machines $ITM[t]$  each working on the population $X[t]$ in generations $t = 0, 1, 2, ...$.
\end{Definition}  

Simple inductive Turing machines are abstract automata (models of algorithms) closest to Turing machines. The difference between simple inductive Turing machines and Turing machines is that a Turing machine for recursive languages always gives the final result after a finite number of steps and after this it stops the process of computation or, at least, the machine informs when the result is obtained. There are different ways to inform that the final result is obtained. For instance, it is possible to have a special symbol in the output alphabet. This symbol is used only to indicate that what is in the output tape is the final result. Thus, when a Turing machine comes to a final state, it repeats the output with this special symbol, indicating that this is the final result. Another way to inform that the final result is obtained is to halt after obtaining this result. It is always possible to assume that after obtaining the final result, the Turing machine stops \cite{hopcroft01}. When starting with some input $x$, a Turing machine never comes to a final state, it does not give its final result for this input.

In a similar way, inductive Turing machines give the final result after a finite number of steps. However, in contrast to Turing machines, inductive Turing machines do not always stop the process of computation or inform when the final result is obtained. In some cases, they do this, while in other cases they continue their computation and give the final result. Namely, when the content of the output tape of a simple inductive Turing machine forever stops changing, it is the final result.

\begin{Remark} Note that it is also possible to consider {\em Reconfigurable evolutionary Turing machines, Weighted evolutionary Turing machines, Evolutionary limit Turing machines, Evolutionary bounded Turing machines, Evolutionary periodic Turing machines and Evolutionary recursively generated Turing machines} \cite{burebe09,burebe10,burebe12,burebe13}. In addition, it is possible to introduce other classes of evolutionary automata, for example, {\em Evolutionary pushdown automata, Evolutionary timed automata} or {\em Evolutionary context-free grammars}, or {\em Evolutionary Chomsky-0 grammars}, and to study relations between known classes of conventional automata and newly introduced classes of evolutionary automata.
\end{Remark}   

\section{Expressiveness of Evolutionary Automata}

Terminal behavior of the evolutionary automaton $E$ defines the {\em terminal language} accepted by this automaton.

\begin{Definition}[On acceptation in the terminal mode]
A population $X$ is accepted in the terminal mode of the evolutionary automaton $E$ if given the population $X$ as input to the evolutionary automaton $E$, there is a number $t = 0, 1, 2, ...$ such that the component automaton $A[t]$ from $E$ comes to an accepting state.
\end{Definition}

\begin{Remark} Note that there are other useful modes of accepting a word by an automaton. For instance, in the infinite mode, the automaton accepts a word $w$ when infinitely many components of $E$ accept this word \cite{hopcroft01} [25].
\end{Remark}

\begin{Definition}[On terminal languages] 
The {\em terminal language TLE} of the automaton $E$ is the set of all words accepted in the terminal mode of the automaton $E$.
\end{Definition}

\begin{Definition} [On local langugaes]
{\em Local languages} (defined by one generation only) are special cases of terminal languages. It is not surprising that local languages are less expressive than terminal ones, because when we consider only one level (one generation) of evolutionary automaton, thus such automaton cannot be more expressive than the evolutionary automata using all its components.
\end{Definition}

\begin{Remark}
However, what is surprising and dissimilar to non-evolutionary classical automata, EFA operating in the terminal mode can be more expressive not only in comparison with finite automata but also in comparison with Turing machines and inductive Turing machines. 
\end{Remark}

For instance, evolutionary finite automata (EFA) can accept arbitrary enumerable languages. The following theorems illustrate how EFAs working in terminal mode can accept context-free or context-sensitive languages that are not acceptable by finite automata \cite{burebe12,ebebur09}.

\begin{Theorem} [On EFA accepting context-free languages]
TL(EFA) contains some context-free languages in the alphabet $\Sigma$.

{\em Proof:} Let us consider an EFA $E$ accepting the context-free language $\{a^nb^n \;|\; n \geq 0\}$. We take a finite automaton $A[0]$ that takes as input population of size one $X[0]$ that is equal to any input string, and  accepts only the empty word $\varepsilon$ and outputs any other word as $X[1]$, a finite automaton $A[1]$ that accepts only the word $ab$ and outputs any other word as $X[2]$, a finite automaton $A[2]$ that accepts only the word $a^2b^2$ and outputs any other word as $X[3]$ and so on. It is easy to see that the context-free language $\{a^nb^n \;|\; n \geq 0\}$ is the terminal language of the evolutionary finite automaton $E$. $\Box$
\end{Theorem}

To help the reader intuition, let us consider context-sensitive languages.

\begin{Theorem}[On EFA accepting contex-sensitive languages]
TL(EFA) contains some context-sensitive languages in the alphabet $\Sigma$.

{\em Proof:} Let us consider an EFA $E$ accepting the context-sensitive language $\{a^nb^nc^n \;|\; n \geq 0\}$. We take a finite automaton $A[0]$ that takes as input population of size one $X[0]$ that is equal to any input string, and  accepts only the empty word $\varepsilon$ and outputs any other word as $X[1]$, a finite automaton $A[1]$ that accepts only the word $abc$ and outputs any other word as $X[2]$, a finite automaton $A[2]$ that accepts only the word 
$a^2b^2c^2$ and outputs any other word as $X[3]$ and so on. 
It is easy to see that the context-sensitive language $\{a^nb^nc^n \;|\; n \geq 0\}$ is the language of the evolutionary finite automaton $E$. $\Box$
\end{Theorem}

\begin{Remark}  In proofs of both Theorems 5.1. and 5.2., we have obviously  finite automata $A[t], t=0, 1, 2,...$ accepting appropriate single strings (we did that for simplicity of the proof only, and finite automata components can accept the set of strings). In infinity, they accept in the terminal mode as the union set the appropriate context-free and context-sensitive languages. Because the finite automata from both theorems have very simple structure, it is quite easily to mutate component finite automata from generation to generation to construct appropriate EFA accepting all the strings from the language. As the input population $X[0]$ we use strings which travel through component finite automata until they find an accepting automaton in the terminal mode. If string belongs to the language the EFA will stop/accept, otherwise will not. 
\end{Remark}

Now, we are ready to present the main result on expressiveness of evolutionary automata.
It would be natural to ask whether evolutionary automata accept recursively enumerable languages as Turing machines do, however a stronger result has been obtained:

\begin{Theorem} ({\sc On acceptance of any terminal language over a given alphabet})\\
TL(EFA) coincides with the class of all languages in the alphabet $\Sigma$.

{\em Proof:} The justification is somehow similar to proofs of Theorems 5.1 and 5.2. We show that given a formal language $L$, i.e., a set of finite words in the alphabet $\Sigma$, there is an EFA $A$ such that $A$ accepts $L$. To do this, for each word $w$, we build a finite automaton $A_w$ that given a word $w$ as its input, accepts only the word $w$, and given any other word $u$, it outputs $u$, which goes as input to the next finite automaton in the EFA $A$. In both cases, the automaton $A$ comes to a terminal state pending that there exists $A_u$ accepting $u$, otherwise $u$ is rejected. Then taking any sequence $E = \{A[t] = A_w, \;w \in  L\}$ of such automata, we obtain the necessary evolutionary finite automaton $A$. $\Box$
\end{Theorem}

Firstly, we can conclude that EFA accept not some but all context-free and context-sensitive languages over a given alphabet.

\begin{Corollary} ({\sc On acceptance of all regular, context-free and context-sensitive languages over a given alphabet}) \\
EFA accept in terminal mode all regular, context-free and context-sensitive langauges 

{\em Proof:} is obvious and follows directly as conclusion from Theorem 5.3 that evolutionary finite automata can accept not some, but all regular, all context-free and all context-sensitive languages. $\Box$
\end{Corollary}

\begin{Remark}
Note that EFAs can accept all languages over alphabet $\Sigma$,
both decidable and undecidable. The justification is vey simple: we repeat generally the same what did Alan Turing, who proved that infinitely many but enumerable number of Turing machines cannot compute all real numbers \cite{turing36} that are equal to all possible subsets of natural numbers as their inputs. However, surprisingly, our conclusions are different - EFA are MORE EXPRESSIVE than Turing machine. Namely, we have an enumerable number of finite automata components possible - each accepting one natural number. All possible EFA (being equal to all subsets of finite automata components) represent all possible languages $L$ over alphabet $\Sigma$. Each EFA represents one possible language. All EFA can compute all real numbers (the difference with Turing's proof - Evolutionary Finite Automata are more expressive than Turing machine) and explanation of the enormous expressiveness of EFA. 

The reader may ask, where is the catch in our reasoning? The problem is how to construct a sequence of finite automata components accepting a given language $L$ in the terminal mode. If we know all the strings of the language, then to construct corresponding finite automata components is straightforward. For undecidable non-recursive languages (e.g., language of the universal TM or diagonalization language \cite{hopcroft01}), we cannot construct such finite automata, thus we will not know that this specific EFA encodes the language of the universal Turing machine, or another undecidable language.  Replacing EFA by more powerful component automata, e.g., pushdown automata or Turing machines will not help either.
\end{Remark}

\begin{Remark} It follows from the above, that expressive power of EC using an infinite (unbounded) number of generations is enormous, and the expressiveness of EFA is the same as expressiveness of EPDA, ETM, EITM. On the other hand, the power of EC using a finite number of generations is consistent with the classical abstract automata theory \cite{hopcroft01}.
\end{Remark}

\section{Expressiveness of Evolutionary Automata versus Other Models of Computation}

The aim of this section is to compare and to relate expressiveness of evolutionary automata and evolutionary computation to other super-Turing models of computation. For obvious reasons, we will not compare evolutionary automata with models of computation equal or less powerful than Turing machines. Simply, they subsume expressiveness of classical ordinary models of computation (see appropriate theorems from previous section). For these reasons, we will concentrate on more expressive super-Turing models of computation.

\begin{Definition}[On super-Turing computation]
By {\em super-Turing computation} (also called {\em hypercomputation}) we mean any computation that cannot be carried out by a Turing machine as well as any (algorithmic) computation carried out by a Turing machine.
\end{Definition}

Super-Turing models derive their higher than the TM expressiveness using three principles interaction, evolution, or infinity:
\begin{itemize}
\item
In the interaction principle the model becomes open and the agent interacts with either a more expressive component or with infinitely many components.
\item
In the evolution principle, the model can evolve to a more expressive one using non-recursive variation operators.
\item
In the infinity principle, models can use unbounded resources: time, memory, the number of computational elements, an unbounded initial configuration, an infinite alphabet, etc.
\end{itemize}

The details can be found in \cite{ebegolweg04,ebeweg03}.

\begin{Example} [A list of some super-Turing models of computation]
In \cite{ebegolweg04,syropoulos08} several super-Turing models have been discussed and overviewed. An {\em incomplete} list includes: 
\begin{itemize}
\item
{\em Turing's o-machines, c-machines and u-machines} (Turing, A.) - they use help of Oracle (o-machines) or human operator (c-machines), or they form an unorganized network that may evolve by genetic algorithms or reinforcement learning (u-machines),
\item
{\em Cellular automata} (von Neumann, J.) - an infinite number of discrete finite automata cells in a regular grid, 
\item
{\em Discrete and analog neural networks} (Garzon, M., Siegelmann, H.) - a potentially infinite number of discrete neurons or neurons with true real-valued inputs/outputs,
\item
{\em Interaction Machines} (Wegner, P.) - they interact with other machines sequentially or in parallel by infinite multiple streams of inputs and outputs, 
\item
{\em Persistent Turing Machines} (Goldin, D.) - they preserve contents of memory tape from computation to computation, 
\item
{\em Site and Internet Machines} (van Leeuwen, J., Wiedermann, J.) - they have input/output ports that allow to interact with an environment or Oracle and communicate by infinite streams of messages, 
\item
{\em The $\pi$-calculus} (Milner, R.) - potentially an infinite number of agents interacting in parallel by message-passing, 
\item
{\em The \$-calculus} (Eberbach, E.) - potentially an infinite number of agents interacting in parallel by message-passing and searching for solutions by built-in $k\Omega$-optimization meta-search that may evolve, 
\item
{\em Inductive Turing Machines} (Burgin, M.) - they may continue computation after providing the results in a finite time,
\item 
{\em Infinite Time Turing Machines} (Hamkins, J.D.) - they allow an infinite number of computational steps,
\item
{\em Accelerating Turing Machines} (Copeland, B.J.) - each instruction requires a half of the time of its predecessor's time forming a geometric convergent series.
\item
{\em Evolutionary Turing Machines} (Eberbach, E.) and {\em Evolutionary Automata} (Burgin M., Eberbach E.) - they allow for evolution of evolution and may use an infinite number of generations. 
\end{itemize}
\end{Example}

 From \cite{wegebebur12} we get.

\begin{Theorem} Terminal languages of Interaction Machines coincide with the class of all languages in the alphabet $\Sigma$. $\Box$
\end{Theorem}

We can safely assume that models based on Oracles, i.e., Turing o-machines \cite{turing39} and Site and Internet Machines can also accept arbitrary languages over a given alphabet.

\begin{Theorem} Terminal languages of o-machines and Site and Internet Machines coincide with the class of all languages in the alphabet $\Sigma$. $\Box$
\end{Theorem}

We believe that analogous proofs can be derived for \$-calculus \cite{ebe07,ebe22}, $\pi$-calculus (pending that replication operator allows for infinity), cellular automata (extended to random automata networks, where each cell may represent a different finite state automaton), neural networks, Turing u-machines \cite{turing48} (pending that they allow for an infinite number of nodes), i.e., models where we can derive the sequence of components inheriting all needed information from their predecessors, i.e., we can repeat essentially the proofs for evolutionary automata and interaction machines. Thus, we will write, based on \cite{ebe22}, the following.  

\begin{Conjecture} Terminal languages for \$-calculus, $\pi$-calculus, cellular automata generalized to random automata networks, neural networks and Turing u-machines coincide with the class of all languages in the alphabet $\Sigma$. $\Box$
\end{Conjecture}

From Theorems 5.3, 6.1 and 6.2 and Conjecture 6.3 we can derive immediately Corollary 6.4.

\begin{Corollary} Expressiveness of \$-calculus, $\pi$-calculus, cellular automata, neural networks, Turing o-machines and u-machines, Evolutionary Automata and Interaction Machines is the same and allow to accept all languages over a given finite alphabet. $\Box$
\end{Corollary}

It is not clear at this moment how to classify expressiveness of Infinite Time Turing Machines and Accelerating Turing Machines - simply, the conditions of an infinite number of steps or doubling the speed of each successive step alone seem not be sufficient to prove that those models can accept all languages over a given alphabet. Similarly, we do not have enough details on c-machines, because Turing did not provide sufficient details about them \cite{turing36}. Also, we cannot properly classify at this moment the expressiveness of Inductive Turing Machines and Persistent Turing Machines. These are left as the open future research problems.
However, taking into account the above, we can write safely at this moment that they can at most match expressiveness of Evolutionary automata, because nothing can exceed that limit.

\section{Computational Complexity of Evolutionary Automata}

Note that to speak properly about computational complexity of evolutionary automata, not a new section only, but several papers would be required. This is so, because an extension of classical computational complexity \cite{hopcroft01,kleinberg06} for infinite cases would be needed. Following, for instance \cite{ebe21}, we considered four time/space complexities classes of algorithms:

\bigskip
\begin{Definition}(On time complexity)
\newline
We can define four classes of problems/
languages for abstract automata/ machines over discrete finite alphabet $\Sigma$ of at least one symbol to decide strings in the language in the order of increasing hardness and complexity. We will define time complexity counting the number of steps, however an analogous definition for space complexity counting the number of memory cells can be defined:

\begin{enumerate}
{\em
\item
{\bf p-decidable problems:} The number of steps is polynomial in the problem size and problems will be called 
{\em polynomially decidable}\\
({\em p-decidable}).
\item
{\bf e-decidable problems:} The number of steps is at least exponential  in the problem size and problems will be called {\em exponentially decidable}
({\em e-decidable}).
\item
{\bf a-decidable problems:} The number of steps is infinite but computed in finite time, i.e., {\em asymptotically/limit
decidable} ({\em a-decidable})
(analogy: convergent infinite series,
mathematical induction, computing infinite sum in definite integral).
\item
{\bf i-decidable problems:} The number of steps is infinite and requires infinite time
to decide strings in the problem size, i.e., {\em infinitely decidable} ({\em i-decidable}) (undecidable in the finite sense).
}
\end{enumerate}
\end{Definition}

The classical complexity theory usually covers classes (1) as easy/tractable and class (2) as intractable problems.
In class (1) polynomials have usually small constant coefficients and exponents. In class (2) functions can be transcomputationally complex, e.g., $1.5^n$, $2^n$, $n!$, $2^{2^n}$, $2^{2^{2^n}}$. Separation of class (1) and (2) is not rigid. There can be complexities between polynomials and exponentials, e.g., $n^{log_2 n}$.

Class (3) is
an intermediate class because although technically it requires an infinite number of steps
(or memory cells for space complexity), we can find
a solution in the limit using finite resources. These can be for instances of infinite geometric series where sum is convergent in infinity and computed by formulas in finite number of steps. The same is with infinite series used to compute numbers like $e$ or $\pi$. The mathematical induction proofs also have an infinite number of steps, however an induction step allows to fold them to finite computations. Class (3) is represented by convergent in infinity subset of
Inductive Turing Machines,
anytime algorithms,
evolutionary algorithms, or \$-calculus.
Class (4) requires infinite resources and is unsolvable by Turing Machine
(but solvable by hypercomputers using infinite resources). Classes (2) , (3) and (4) cover non-polynomial algorithms,
classes (3) and (4)
belong to super-recursive algorithms \cite{burgin05}. Class (1) and (2) cover recursive algorithms.

Classes (1) and (2) cover recursive algorithms (polynomial and intractable problems). Classes (3) and (4) go outside classical computation complexity and cover super-recursive algorithms (undecidable problems) some of them convergent in the limit/infinity and some not.

Undecidable problems are characterized by computations growing faster than exponentially (i.e., hyper-exponentially) or they may have even an infinite computational complexity (e.g., an enumerable cardinality like for natural numbers (positive integers), or real type like for real numbers). 
Undecidable problems DO HAVE their computational complexity - we do not have good theory to catch properly that yet.

Note that we have in reality an infinite hierarchy of infinities/cardinalities in mathematics/set theory \cite{kuratowski77}, whereas computer science considers only typically enumerable infinity (denoted by omega $\omega$ , alpha $\alpha$  or  aleph 0 $\aleph_0$ ) with some extension to true real numbers (denoted by $c$ or aleph 1 $\aleph_1$) represented by analog computers, neural networks operating on real numbers or evolution strategies operating on vectors of real numbers.

Traditional complexity theory deals with various type of decidable algorithms (see e.g., \cite{hopcroft01,kleinberg06}), i.e., graph, greedy, divide and conquer, dynamic programming, network flow, NP-complete, PSpace, approximation, local search, and randomized algorithms. All of them have polynomial or exponential complexities. Algorithms than run forever \cite{kleinberg06} and super-recursive algorithms \cite{burgin99,burgin05} require an extension of complexity theory to infinite cases (nobody did it so far for enumerable $\aleph_0$ or non-enumerable infinities $\aleph_1$, $\aleph_2$, $\aleph_3$, and so on. As we wrote in section 2 evolutionary algorithms are a hybrid class having features both of local search, approximation, randomized, parallel and running forever algorithms \cite{kleinberg06}. They allow to decrease an exponential complexity for hard optimization problems, but for the price of reaching local optima only. This means that local search, also known as hill climbing, typically allows to find local optima only. However, evolutionary algorithms if the search is complete and uses elitist selection allows in infinity to reach the global optimum of the fitness function (as we wrote in section 3). Then, we do not solve a given problem in polynomial time nor in an exponential time, but, perhaps, in infinite time (classes (3) and (4) of algorithms). But if infinite, what type of infinity we are talking about, $\aleph_0$  or $\aleph_1$?  Or something else?

Evolutionary automata in each generation (local mode) share complexities of conventional evolutionary algorithms. If we limit the number of generations to a finite number, the complexity of a local mode (a single generation) and terminal mode (all generations) will be the same.

\section{Conclusions and Final Comments}

In general, Evolutionary Automata (EA) do not have to take one word from language $L$ and to construct for it recognizing Finite Automaton - this was done intentionally to simplify the proof of expressiveness of EA. Evolutionary automata components do not need to be of class of finite automata at all. However, for illustration, let's restrict our attention to EFA.  If EFA take as input undecidable languages like language of Universal TM (recursively enumerable but not recursive) or diagonalization language (non-recursively enumerable), we should be able to construct such input strings belonging to those languages (not done yet - it requires to solve the membership problem for undecidable languages), and next to construct corresponding finite automaton component (that step is trivial), and, finally, to decide in final number of generations/steps such input string. For decidable languages, this will be easier, because the membership problem for their corresponding strings has been solved. Note, that we do not provide string encodings for undeciable languages - we prove only that such encoding exists (look, e.g.,  \cite{hopcroft01}).  

Normally, the process of evolution in Evolutionary Algorithms is forced to stop after a finite number of generations (for the price that perhaps we will miss a better/fitter solution). And there DEEP Evolutionary Computation with Evolutionary Automata model are in play to help, with a larger/deeper number of generations (potentially infinitely many), providing the BENEFIT of developing globally optimal solutions with the best fitness value. This corresponds directly to adding more hidden layers in Deep Neural Networks that caused current boom and success of Artificial Intelligence and Machine Learning. According to appropriate theorems for NNs, we need only 3 layers for NNs to express any mapping from real numbers to real numbers, thus Deep NNs should not benefit from adding more hidden layers (but they obviously did BENEFIT for many applications). Similarly, we hope, it may be with Deep Evolutionary Computation by adding deeper (having more generations) evolution could be BENEFICIAL too. 

Another approach to Deep Evolutionary Computation represent Evolutionary Artificial Neural Networks (EANN) \cite{fogel02}. EANN are neural networks where weights are modified by evolutionary algorithms and not by backpropagation. There, like in other types of Deep Neural Networks, the number of EANN hidden layers can be increased.

This paper demonstrates the potential tremendous expressiveness of evolutionary computation and natural evolution (assuming that natural evolution is not restricted to our planet only, i.e., applies to the whole Universe). At this moment nobody in the world proved truth or falsity of two opposite scientific hypotheses, i.e., whether the real world is finite or infinite. In this paper the proof of such tremendous expressiveness has been provided for evolutionary automata, thus it is reasonable to hypothesize and extrapolate that result to natural evolution too. 

The arguments for expressiveness of evolutionary computation and natural evolution are objective. In the paper it was never claimed that models of computation as powerful as or less powerful than Turing machine are not interesting. The whole classical evolutionary computation uses commonly such models. Everything what is claimed is that it is worth to look at models more expressive than Turing machine too. If evolutionary automata are so expressive, then it is worth to use them in evolutionary computation. Nowhere in the paper is claimed that evolution in nature requires Turing computability or more. This was exactly left for discussion. Everything what is claimed is that because evolutionary automata may use an infinite number of generations and self-evolution that natural evolution uses as well, then perhaps natural evolution is similarly expressive like evolutionary automata. This is a perfectly valid and reasonable another scientific research hypothesis (that cannot be proven without formal definition of natural evolution).
This by itself should be very interesting for general readers and specialists in the area. We hope that this will stir a fruitful discussion and comments from the readers. This is reasonable because the whole mathematics and computer science is based and uses commonly infinities of various types. The same applies to probabilistic hill climbing evolutionary computation, because probabilities are based on infinite samples in mutation, crossover, selection operators, Markov chain analysis, schema theorems, ergodic convergence - all of them are based and use infinities of various types. 

The results from this paper were possible due to introduction of the evolutionary automata model that allows evolution of evolutionary algorithms and to use an infinite number of generations. Note that if to use fixed evolutionary algorithms with a finite number of generations we are back in the class of current evolutionary algorithms.

We stressed in the paper that evolutionary algorithms form a hybrid and a very rich class with features of both local search, approximation, randomized, parallel and running forever algorithms \cite{kleinberg06,michfog04}.

Of course, we are aware that much more research is needed. Probably, some preliminary ideas from this paper will be rejected after careful verification and many new unexpected results will follow. The practical implications to the EC field, suggestions and guidelines of this study to the EC researchers and practitioners is that it is worth to consider an infinite number of generations and to evolve evolutionary algorithms themselves, because this allows to extend the class of problems to be solved, including finding global optima in infinity, and to approximate solutions of TM undecidable problems.

Note, that it took almost 40 years for the Fifth Generation Computing to result finally in the booming AI after many successes and failures. Now the trends are opposite, whether we should restrict AI, because it became too powerful. Similar situation has happened with the Sixth Generation Computing that led with big struggling and zigzagging to Deep Neural Networks. We can expect that the same, most likely, will happen with biomolecular computing (Seventh Generation Computing) and super-Turing computing too. To overcome such new limits, we need urgently the breakthrough results from Undecidability/Incomputability Theorists, because unfortunately we cannot count in this aspect anymore on help from Kurt G\"{o}del, Alonzo Church, Alan Turing or John von Neumann \cite{ebemik14}. Would Alan Turing be happy or disappointed if he could learn that his dream about AI \cite{turing48} has finally materialized to such degree that people stopped to doubt whether AI was possible, but instead, they became afraid of unrestricted power of AI?

\subsection*{Acknowledgements}
The author would like to thank Prof. Philip C. Treleaven for introducing him to the Fifth, Sixth and Seventh Generation Computing and new programming paradigms and architectures while at University College London. The author is grateful to late Prof. Peter Wegner from Brown University for multiple discussions on completeness of Turing computability, his Interaction Machines and fruitful cooperation. Finally, late Prof. Mark Burgin from UCLA must be acknowledged for his super-recursive algorithms, Inductive Turing Machines, and multiple contributions to the theory of evolutionary automata. 
The author appreciates useful comments from anonymous reviewers and editors.

\end{document}